# An Improved Positioning Accuracy Method of a Robot Based on Particle Filter

**Rashid Ali, Dil Nawaz Hakro, Yongping He, Wenpeng Fu, and Zhiqiang Cao**

**Abstract** This paper aims to improve the performance and positioning accuracy of a robot by using the particle filter method. The laser range information is a wireless navigation system mainly used to measure, position, and control autonomous robots. Its localization is more flexible to control than wired guidance systems. However, the navigation through the laser range finder occurs with a large positioning error while it moves or turns fast. For solving this problem, the paper proposes a method to improve the positioning accuracy of a robot in an indoor environment by using a particle filter with robust characteristics in a nonlinear or non-Gaussian system. In this experiment, a robot is equipped with a laser range finder, two encoders, and a gyro for navigation to verify the positioning accuracy and performance. The positioning accuracy and performance could improve by approximately 85.5% in this proposed method.

## 1 Introduction

The localization measurement technology is a technique for measuring a location using physical information, geographical information, and logical information of a moving object, which is primarily divided into local localization and global localization. Local localization measurement mainly refers to measuring a moving object's

R. Ali · Y. He · W. Fu · Z. Cao
School of Information Engineering, Southwest University of Science and Technology, Mianyang 621010, China
e-mail: rashidcs@uot.edu.pk

R. Ali
Department of Computer Science, University of Turbat, Turbat 92600, Pakistan

D. N. Hakro
Institute of Information Technology, University of Sindh, Jamshoro, Pakistan

relative coordinates using an encoder and a gyro [1]. There is a disadvantage in that the reaction speed is fast and resistant to disturbance, but the accuracy of the position measurement becomes low due to the accumulation of errors. On the other hand, global positioning measurement is a method to measure the position in global coordinates using pure differential global positioning system (DGPS), LRF, ultrasonic satellite, and radio frequency identification (RFID). The measurement error is predictable, and there is no error accumulation problem. However, the reaction speed is slower than that of a local localization measuring device, and several meters to tens of meters occur depending on the indoor environment using EKF to fuse information from optical wheel encoders, an accelerometer, and a gyroscope [2].

As a device for measuring the position using information, the installed reflector's position must be known in advance. In more detail, the LRF header rotates and transmits the laser. The laser reflects and returns to the reflector, measures the reflector's distance, and calculates the position and direction by matching the known reflector information. It is more robust and more resistant to disturbances than other global positioning devices using a high-precision laser. Besides, there is an advantage in that the working space can be easily expanded and changed by adjusting the position and number of reflectors. However, due to the header rotating characteristics and the laser to recognize the reflector, errors in rotation and high-speed driving are large [3]. There is a problem in that an object having a high reflectance is recognized as a reflector. Besides, when the position on the target rotation center axis is different from the position of the LRF when applied to the driving device, due to an issue, the position error is amplified in the coordinate conversion process to correct it. Accordingly, research has been actively conducted to correct reflector matching and position measurement errors through probabilistic techniques to solve navigation problems [4]. A representative probabilistic approach is sensor fusion technology through a Kalman filter. However, this study mainly compensated for other sensors using LRF as an observer and correcting only the white noise system.

There are reflector matching problems such as laser navigation, the measurement error problem during rotation, and high-speed driving, which could not be solved. So, for minimizing the problem, the improved positioning accuracy technique has been used [5, 6]. In this paper, we propose a method to improve the accuracy of position measurement using particle filters with strengths in a non-Gaussian system. The proposed method predicts the change in the position and direction of particles distributed in the working space using an encoder and a gyro, which calculates each particle's weight using reflector information received from LRF. It is a method of redistributing particles and estimating the position of the moving object.

## 2 System Configuration

Figure 1a, in this proposed method, an axle driving forklift type LGV robot is used, in which driving and steering are performed at the same position. The local position measurement (LPM) of the LGV robot is measured by using and installing a low-cost

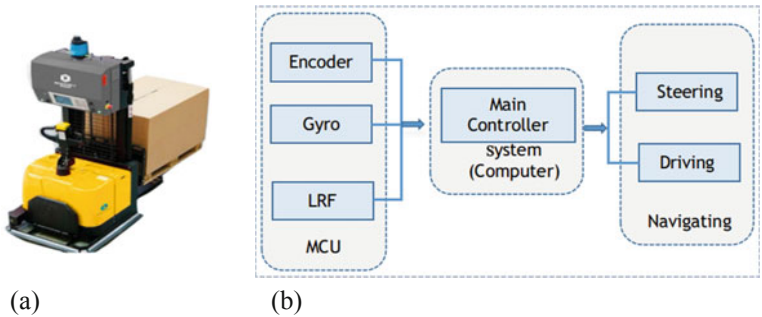

**Fig. 1** **a** Structure of LGV robot and **b** system conf guration

and eff cient vehicle guided encoders and gyros to measure linear velocity and angular velocity and moreover f duciary markers has been used to represent the direction of the vehicle [7, 8]. Also, the SICK LRF is used for the global location measurement. The LRF is installed on the top of the vehicle robot with a height of 1.9 m, which is the highest position in the LGV robot. The laser was minimally affected by the surrounding objects. The encoder is installed on the auxiliary wheel under the fork, and the gyro is installed near the driving wheel.

Figure 1b shows the overall system conf guration of the LGV robot, and the LGV robot system consists of a position measuring unit, a control unit, and a driving unit. The position measurement is performed by using the LRF, gyro, and encoder. The measured information is transmitted every 100 ms through a microcontroller unit (MCU) sent to the main controller, which calculates the LGV robot's position based on the transmitted information and controls the driving unit.

## 2.1 Sensors Fusion and Analysis

Figure 2 shows the system overview where the LRF (NAV200) transmits a laser while the header rotates 360° and measures the ref ector's distance using the time difference. The current position is also calculated by matching the position information of the measured ref ectors with the position information of the known ref ectors [9]. The encoder used to measure the guided robot's angular and linear velocities is a LIB-49B model, which has a resolution of 1000 pulses, and the gyro my-Gyro-300 SPI model, which has a sensitivity of ±360°/s. The conf gured sensors fused the sensors through particle f lter (FP) for the precise position measurement. A particle f lter (PF) was applied to improve the positional accuracy of laser navigation.

Through sensor fusion, the position measurement system, such as laser range f nder, encoders, and gyro, builds a f exible system; that is robust against disturbances and changes the working environment and work contents [10]. It compares the error between the position measurement result of the laser navigation and the PF position measurement results while installing the sensors on the LGV robot. The particle filter method was applied as a fusion method of the internal sensors and the external sensors.

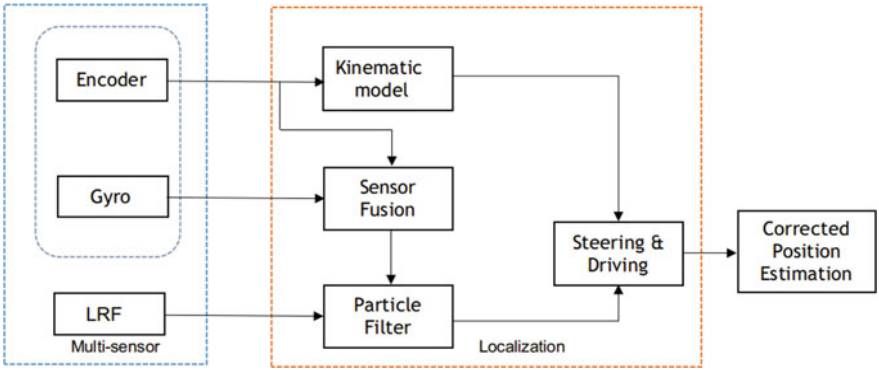

**Fig. 2** System overview

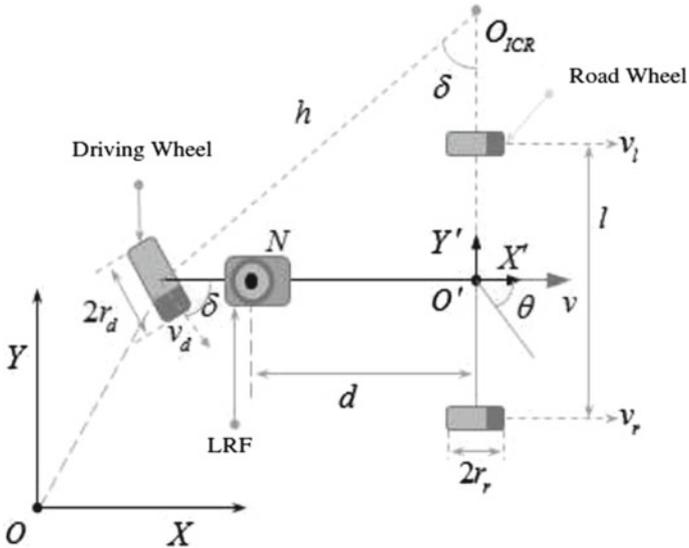

**Fig. 3** Kinematics model of LGV

## 2.2 Kinematics Model Impact on the Robot

Figure 3 shows the laser-guided robot's kinematic model with the axle drive used in the experiment.

In the f gure, $h$ is the distance between the center of rotation $O_{ICR}$ and the center of the driving wheel, the angular velocity $\dot{\delta}$, the linear velocity $v_d$, and the linear velocity $v$ of the driving unit can be calculated through (1).

$$\dot{\delta} = \frac{v_d}{h} = \frac{v}{h\cos\delta}, v_d, w_d, v = v_d \cos\delta \qquad (1)$$

In this paper, two encoders are installed on the auxiliary wheel to measure each wheel's angular velocity $w_l$ and $w_r$ [11]. The linear speed of each wheel is calculated as (2), and the LGV linear speed $v$ and angular speed $\omega$ are calculated as (3).

$$\begin{aligned} v_l &= w_l \times 2r_l \\ v_l &= w_r \times 2r_r \end{aligned} \qquad (2)$$

$$\begin{cases} v = \dfrac{v_r + v_l}{2} \\ w = \tan^{-1}\dfrac{v_r + v_l}{l} \end{cases} \qquad (3)$$

Using (2) and (3), the amount of change in the LGV coordinates and angles can be calculated as follows.

$$\begin{bmatrix} \dot{x} & \dot{y} & \dot{\theta} \end{bmatrix}^T = \begin{bmatrix} v\cos w \\ v\sin w \\ w \end{bmatrix} \qquad (4)$$

The (4) is the displacement of the target rotation center axis $O$. The position measured through laser navigation is $N$. In general, forklift LGV robot controls based on the target rotation center axis [12]. So the position measured in the laser navigation is converted to the target rotation center axis using (5). In (5), POS is the LGV position measured in the laser navigation, where the POS is converted to the target rotation center axis.

$$\begin{aligned} \text{POS} &= \text{POS}_x d\ \cos(\text{POS}_\theta) \\ \text{POS} &= \text{POS}_y d\ \sin(\text{POS}_\theta) \end{aligned} \qquad (5)$$

## 3 Position Measurement Based on Particle Filter

The particle filters are a stochastic approach to Bayesian filters that predict dynamic system conditions from noisy inputs and show high efficiency in nonlinear or non-Gaussian systems [13]. Also, unlike other global positioning techniques, an efficient prediction is possible by focusing on a state in which the likelihood of observation data is high.

### 3.1 Analysis on Bayesian Filter

Figure 4 shows the Bayesian filter estimation process, a widely used technique for recursively estimating a system's state from sensor measurements [14].

Figure 4 where $x_t^-$ means the state of $t$ time predicted at time $t-1$, and $x_t$ is the estimated state of $t$ time. $u_t$ is the control input of the system model, and $z_t$ is the observation data. The posterior probability $x_t$ is the final estimate at time $t$, represented as (6).

$$bel(x_t) = p(x_t | z_t, u_t, z_{t-1}, u_{t-1}, \ldots, z_0, u_0) \tag{6}$$

This can be expressed in two steps:

$$bel^-(x_t) \leftarrow \int p(x_t | x_t^-, u_t) \, Bel(x_t^-) \, dx_t^- \tag{7}$$

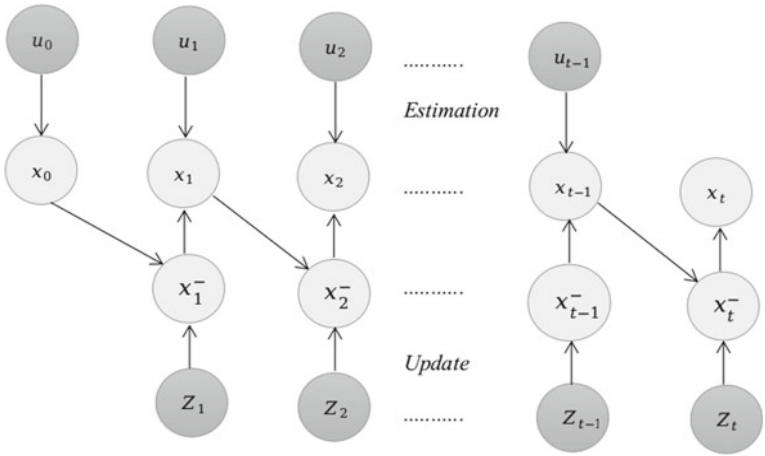

**Fig. 4** Bayes filter

$$bel^-(x_t) \leftarrow \alpha p(z_t|x_t) Bel^-(x_t) \tag{8}$$

The prediction step for predicting the next state estimate from the previous posterior estimate using the system model is expressed as (7). The correction step for correcting the predicted estimate using observation data can be described as (8), while in (8), alpha is a normalization constant [15]. In Eqs. (7) and (8), $p(x_t, x_t^-, u_t)$ represents the system model, and $p(z_t|x_t)$ represents the likelihood of observation data at time $t$.

## 3.2 Proposed Positioning Method

When the laser sensor's measurement position differs from the guided robot target rotational center axis, a coordinate conversion process, as in (4), is required to control the robot effectively. However, since the coordinate transformation is performed using information, including an error in this process, the error may be amplified to control the estimation of the speed [16, 17]. In the proposed method, by setting the coordinates of each particle as the robot's target rotation center axis, such a step is unnecessary so that the coordinate conversion process can be solved where the error amplification problem has been generated.

The proposed positioning method's process is divided into initialization, prediction, correction, and redistribution steps. Initialization is the step of setting the initial position of particles throughout the work environment. If the number of particles is small in the localization, it is highly likely to converge to the local minimum, which is unsuitable for real-time systems because it takes too long. Therefore, in the proposed method, the number of particles was set to 150, which is the most efficient for location measurement precision and calculation time through experiments and evenly distributed throughout the working environment.

In the prediction stage, the particle distribution information at time $t-1$ and the control input data measured from the encoder and gyro sensor are applied to (9) based on kinematics to predict the robot's state change at time $t$.

$$\begin{bmatrix} x_t \\ y_t \\ \theta_t \end{bmatrix} = \begin{bmatrix} x_{t-1} + v_t \cos(\theta_{t-1} + w_t) \\ y_{t-1} + v_t \sin(\theta_{t-1} + w_t) \\ \theta_{t-1} + w_t \end{bmatrix} \tag{9}$$

Figure 5a and b, after the prediction step [18], the reflector's position measured by the laser navigation, is converted to a Cartesian coordinate system at the particle position to correct the system state. The distance difference as reflector 1, reflector 2, and reflector 3, from the known reflectors and LRF, which usually match the detected and measured reflector marks within its working environment, is calculated.

In Fig. 5b, the gray that shows the measured reflector and the black circle represent the actual reflector's information. The closer the distance difference from the actual ref ector, the more likely the real vehicle robot is located on the particle as it approaches zero. Therefore, to obtain the weight according to the distance value, the sum of the distance values is applied to the standard normal distribution with a mean of 0 and a deviation of 1, and the particles' weight is calculated in (10). In the equation, $z_t$ and $\hat{z}_t$ are each real ref ector's positions and the measured ref ector.

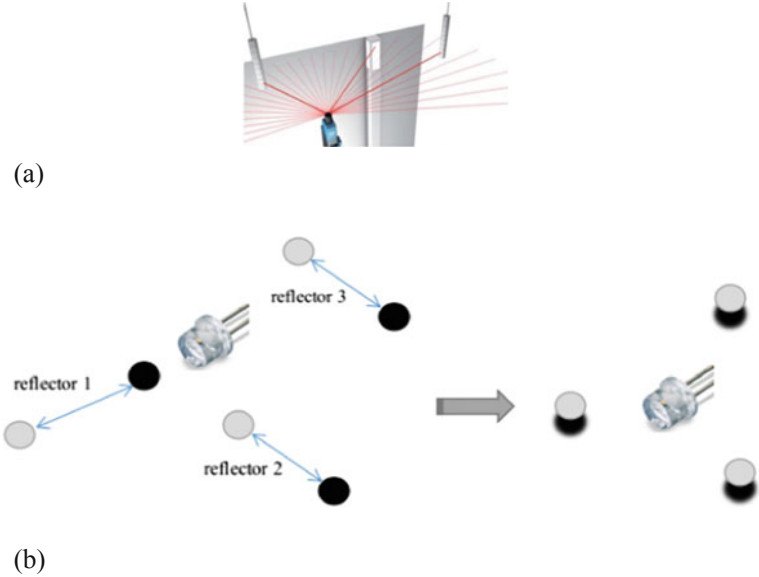

(a)

(b)

**Fig. 5** **a** Laser and ref ectors ref ection and **b** laser guided robot state estimation process

$$w^t = \det(2\pi\ )^{-1/2} \exp\left(-1/2\ (\hat{z}_t - z_t)^T\ \ ^{-1}\ (\hat{z}_t - z_t)\right) \quad (10)$$

To correct the current state using each particle's weight, the corrected current state using a weighted mean is strong in disturbance and has a low error probability. As shown in (11), a system with $M$ particles represented the weighted average calculation.

$$x_t = \frac{\sum_{m=1}^{M} w_t^m x_t^m}{\sum_{m=1}^{M} w_t^m} \quad (11)$$

After calculating the weighted average, a redistribution process is performed to generate a new set of particles to be used in the next prediction step [19]. The redistribution process duplicates or deletes particles according to their weights to converge to an optimal solution. If the particles' weight is high, it means that the robot is likely to be located at the corresponding location. Therefore, in the proposed method, 95% of particles are distributed in the periphery of the top 25% or more particles by experiment. The range is set to ±25 cm when the particles are distributed.

The robot's driving speed in the experiment is 36 cm/s. Since the laser navigation measures data every 450 ms, the robot can move the maximum distance during the time during which the laser navigation measures data is 17 cm. Therefore, considering this value and the measurement error, the redistribution range of particles was set to ±25 cm to rule out the possibility of not converging the optimal solution. Particles below 15% are evenly distributed throughout the working environment. A randomly generated particle throughout the working environment is to rule out the possibility of converging local minimums [20]. After the correction and redistribution step, the robot's position recursively estimated from the redistribution step to the prediction step to estimate the next state.

## 4 Experimentation and Results

Figure 6a and Table 1 show the results of the RMSE and error variance of the entire experiment results and the trajectory rotation of one of the entire experiments. In this experiment, the reflector was used to maximize the reflectivity. Twenty reflectors were installed on the wall by attaching it to a cylinder with a diameter of 10 cm and a height of 85 cm. The size of the environment where the reflector is installed is 25 m × 20 m, and the size of the experimental space used is mainly 9.4 m × 12 m. As for the experimental method, the positioning accuracy was measured during eight repetitive rotations while the robot driving speed and steering angle were fixed at 36 cm/s and 60°, respectively. Besides, to compare and analyze the proposed method's performance [21], it was compared with the results of the position measurement of the laser navigation system.

The RMSE average result in millimeter (mm) of the proposed method and sensors showed 20.4341 mm and 56.8978 mm, respectively, and the dispersion average of 117.6301 mm and 472.6535 mm, respectively. As shown in Table 1, it can be confirmed that the accuracy of the position measurement of the proposed method is high.

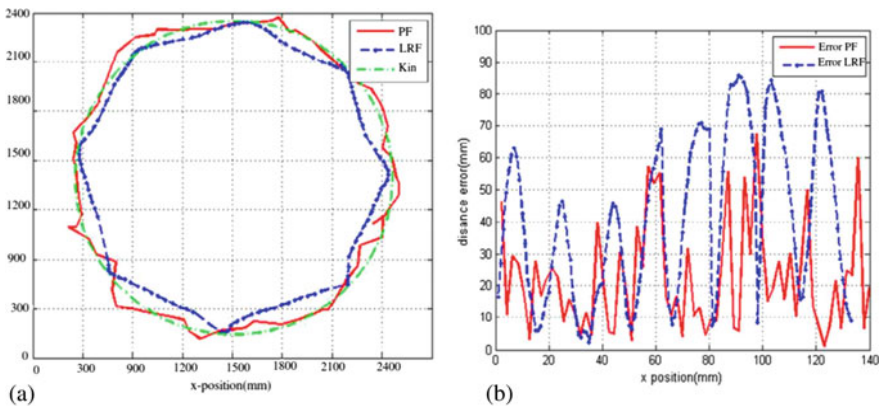

**Fig. 6** **a** Positioning result and **b** results of positioning error

**Table 1** Comparison of variance and RMSE

| Number | Navigation type (laser-guided) | | Method (particle filter) | |
|---|---|---|---|---|
| | RMSE | Variance ($\sigma^2$) | RMSE | Variance ($\sigma^2$) |
| 1 | 56.6685 | 434.2556 | 20.2012 | 81.7710 |
| 2 | 52.3040 | 441.1461 | 13.5170 | 151.4454 |
| 3 | 56.5035 | 417.0178 | 19.7110 | 71.7040 |
| 4 | 54.1946 | 413.6437 | 20.1831 | 206.4672 |
| 5 | 56.8387 | 621.2341 | 17.1157 | 75.8677 |
| 6 | 61.3273 | 471.1634 | 24.3113 | 78.4011 |
| 7 | 56.3223 | 516.4131 | 25.2221 | 165.5830 |
| 8 | 61.0237 | 466.3542 | 23.2112 | 109.8011 |
| Average (mm) | 56.8978 | 472.6535 | 20.4341 | 117.6301 |

Figure 6b shows the result of finding the respective errors by comparing the proposed method and measuring the laser navigation position with the trajectory, calculated using the kinematics model. In the figure, the solid line and the dotted line are the robot's driving trajectories measured using the proposed method and laser sensor, respectively, which is an ideal trajectory while driving. The experiment is better when the steering angle is predetermined, so the shape is closer to the circle. As per the laser sensor results, it is possible to check the part where the shape of the circle is distorted due to many errors in the coordinate conversion process to the target rotation center axis of the robot [5, 21]. On the other hand, the proposed method's result can confirm the shape similar to the circle.

As a result of eight experiments, the errors of 38.8070–50.5045 mm were reduced due to the proposed method and the location measurement of the laser navigation. A performance that improved about 66.5% was confirmed compared to the laser navigation location measurement.

## 5 Conclusion

This paper improves the position measurement performance of laser navigation using the particle filters method. In the conventional laser navigation, there is a problem that a very large error occurs at high speed and rotational driving, and an error is amplified because the position of the sensor and the center axis of rotation of the laser-guided robot do not match. This paper proposed a method to improve the positioning performance of laser-guided navigation using particle filters to solve this problem. In the proposed method, to solve the coordinate transformation problem, each particle is set to the central axis of the robot's rotation target. Also, kinematics were applied to the robot's angular and linear velocities measured by a gyro and an encoder to predict the change in each particle's position and direction.

Moreover, an observation model was designed to determine the weight of particles and finally estimate the robot's position using reflector information measured through laser navigation. To verify the proposed method, the robot that was designed and manufactured by ourselves was used and compared with a commercially available laser navigation location measurement result in a rotational driving experiment in which the positioning error of the laser navigation was large. As a result of the experiment, it was confirmed that the proposed method improved the performance by 85.5% compared to laser navigation. In the future, to improve the performance of this research by considering the current particle distribution and driving conditions with the number of particles and its plans to conduct research to optimize the redistribution scope dynamically.